\documentclass[11pt,a4paper]{article}
\usepackage[hyperref]{acl2020}
\usepackage{times}
\usepackage{latexsym}
\usepackage{amsmath,amssymb}
\usepackage{siunitx}
\usepackage{xspace}
\usepackage{booktabs}
\usepackage{graphicx}
\usepackage{multirow}
\usepackage{xcolor}
\usepackage{subcaption}
\usepackage{microtype}
\newcommand{\fastText}{FastText\xspace}

\aclfinalcopy
\def\aclpaperid{35}

\title{Adversarial Alignment of Multilingual Models \\for Extracting Temporal Expressions from Text}

\author{Lukas Lange$^{1,2,3}$ \\
	\\ \And
	Anastasiia Iurshina$^1$ \\
	\hspace{4cm}$^1$ Bosch Center for Artificial Intelligence, Renningen, Germany\\
	\hspace{4cm}$^2$ Spoken Language Systems (LSV), Saarland University, Saarbr\"{u}cken, Germany\\
	\hspace{4cm}$^3$ Saarbr\"{u}cken Graduate School of Computer Science, Saarbr\"{u}cken, Germany\\
	{\tt \hspace{4cm}\{Lukas.Lange,Heike.Adel,Jannik.Stroetgen\}@de.bosch.com} \\ \And
	Heike Adel$^1$ \\ \And
	Jannik Str\"{o}tgen$^1$ \\
}
\date{}

\begin{document}
\maketitle

\begin{abstract}
    Although temporal tagging is still dominated by rule-based systems, 
    there have been recent attempts at neural temporal taggers. However, all of them focus on monolingual settings. In this paper, we explore multilingual methods for the extraction of temporal expressions from text and investigate adversarial training for aligning embedding spaces to one common space. With this, we create a single multilingual model that can also be transferred to unseen languages and set the new state of the art in those cross-lingual transfer experiments. 
\end{abstract}
\section{Introduction}
The extraction of temporal expressions from text is an important processing step for many applications, such as topic detection and questions answering \cite{StroetgenGertz2016}. 
However, there is a lack of multilingual models for this task. While recent temporal taggers, such as the work by \newcite{laparra-etal-2018-characters} focus on English, only little work was dedicated to multilingual temporal tagging so far.

\newcite{strotgen-gertz-2015-baseline} proposed to automatically generate language resources for the rule-based temporal tagger HeidelTime, but all of these models are language specific and can only process texts from a fixed language. 
In this paper, we propose to overcome this limitation by training a single model on multiple languages to extract temporal expressions from text. 
We experiment with recurrent neural networks using \fastText embeddings~\cite{bojanowski-etal-2017-enriching} and the multilingual version of BERT~\cite{devlin-etal-2019-bert}. 
In order to process multilingual texts, we investigate an unsupervised alignment technique based on adversarial training, making it applicable to zero- or low-resource scenarios and compare it to standard dictionary-based alternatives~\cite{MikolovLS13}.

We demonstrate that it is possible to achieve competitive performance with a single multilingual model trained jointly on English, Spanish and Portuguese.
Further, we demonstrate that this multilingual model can be transferred to new languages, for which the model has not seen any labeled sentences during training by applying it to unseen French, Catalan, Basque, and German data.
Our model shows superior performance compared to HeidelTime~\cite{strotgen-gertz-2015-baseline} and sets new state-of-the-art results in the cross-lingual extraction of temporal expressions.

\begin{figure*}
    \centering
    \includegraphics[width=.99\textwidth]{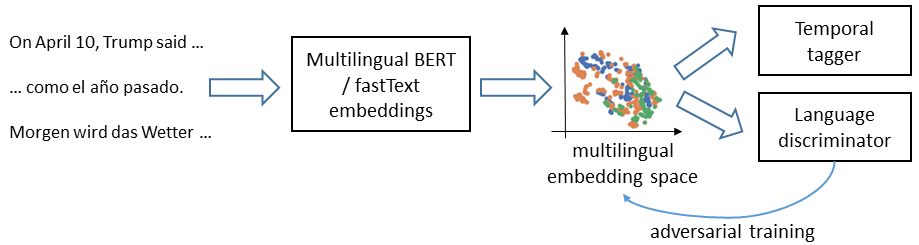}
    \caption{Overview of our multilingual system with adversarial training for improving the embedding space.}
    \label{fig:system}
\end{figure*}

\section{Related Work}
\paragraph{Temporal Tagging.}
The current state of the art for temporal tagging are rule-based systems, such as HeidelTime~\cite{strotgen-gert-2013-journal} or SUTime~\cite{chang-manning-2012-sutime}. 
In particular, HeidelTime uses a different set of rules depending on the language and domain. 
\newcite{strotgen-gertz-2015-baseline} automatically generated HeidelTime rules for more than 200 languages in order to support many languages. However, the quality of these rules does not match the high quality of manually created rules
and the models are still language specific. 
Aside from rule-based systems, \newcite{lee-etal-2014-context} proposed to learn context-dependent semantic parsers for extracting temporal expressions from text.
\newcite{laparra-etal-2018-characters} made a first step towards neural models by using recurrent neural networks.
However, they only performed experiments on English corpora using monolingual models. 
In contrast, we propose a truly multilingual model.

\paragraph{Multilingual Embeddings.}
Recently, it became popular to train embedding models on resources from many languages jointly~\cite{lample2019cross, conneau2019unsupervised}. 
For example, multilingual BERT~\cite{devlin-etal-2019-bert} was trained on Wikipedia articles from more than 100 languages. 
Although performance improvements
show the possibility to use multilingual BERT in monolingual~\cite{hakala-pyysalo-2019-biomedical}, multilingual~\cite{tsai-etal-2019-small} and cross-lingual settings~\cite{wu-dredze-2019-beto}, it has been questioned whether multilingual BERT is truly multilingual~\cite{pires-etal-2019-multilingual, singh-etal-2019-bert, libovicky2019language}. 
Therefore, we will investigate the benefits of aligning its embeddings in our experiments.

\paragraph{Aligning Embedding Spaces.}
A common method to create multilingual embedding spaces is the alignment of monolingual embeddings~\cite{MikolovLS13,joulin-etal-2018-loss}. 
\newcite{smith-etal-2017-offline} proposed to align embedding spaces by creating orthogonal 
transformation matrices based on bilingual dictionaries,
which we use as baseline alignment method.

It was shown that BERT can also benefit from alignment, i.a. in cross-lingual~\cite{schuster-etal-2019-cross, liu-etal-2019-investigating} or multilingual settings \cite{cao-etal-2020-multilingual}. 
In contrast to prior work, we experiment with aligning BERT using adversarial training, which is related to using adversarial training for domain adaptation \cite{ganin-etal-2016-domain}, coping with bias or confounding variables
\cite{li-etal-2018-towards,raff-sylvester-2018-gradient,zhang-etal-2018-bias,barrett-etal-2019-adversarial,mchardy-etal-2019-adversarial}
or transferring models from a source to a target language
\cite{zhang-etal-2017-adversarial,keung-etal-2019-adversarial,wang-etal-2019-weakly}.
Similar to \newcite{chen-cardie-2018-multinomial}, we use a multinomial discriminator in our setting.

\section{Methods}
We model the task of extracting temporal expressions as a sequence tagging problem and explore the performance of state-of-the-art recurrent neural networks with \fastText and BERT embeddings, respectively.
In particular, we train multilingual models that process all languages in the same model. 
To create and improve the multilingual embedding spaces, we propose an unsupervised alignment approach based on adversarial training and compare it to two baseline approaches.
Figure \ref{fig:system} provides an overview of the system. The different components are described in detail in the following.

\subsection{Temporal Expression Extraction Model}
Following previous work, e.g., \newcite{lample-etal-2016-neural}, we train a bidirectional long-short term memory network (BiLSTM) \cite{lstm/Hochreiter97} with a conditional random field (CRF) \cite{lafferty01-crf} output layer. As input, we experiment with two embedding methods: (i) pre-trained \fastText \cite{bojanowski-etal-2017-enriching} word embeddings from multiple languages,\footnote{\url{https://fasttext.cc/docs/en/crawl-vectors.html}} and (ii) multilingual BERT \cite{devlin-etal-2019-bert} embeddings.\footnote{\url{https://github.com/google-research/bert/blob/master/multilingual.md}}
For BERT, we use the averaged output of the last four layers as input to the BiLSTM and fine-tune the whole model during the training of temporal information extraction.
We also experimented with a BERT setup similar to \newcite{devlin-etal-2019-bert} where the embeddings are directly mapped to the label space and the softmax function is used to compute the label probabilities instead of a CRF. However, we found superior performance for the BiLSTM-CRF models.

\subsection{Alignment of Embeddings}
We propose an unsupervised approach based on adversarial training to align multilingual embeddings in a common space (Section \ref{sec:adversarialalignment}) and compare it with two approaches from related work based on linear transformation matrices (Section \ref{sec:baselinealignment}).

\subsubsection{Baseline Alignment}
\label{sec:baselinealignment}
Embedding spaces are typically aligned using a linear transformation based on bilingual dictionaries. 
We follow the work from \newcite{smith-etal-2017-offline}, 
and align embedding spaces based on orthogonal transformation matrices. 
These matrices can either be constructed in an unsupervised way by using words that appear in the vocabularies from both languages, i.e., equal words that can be identified using string matching, or in a supervised way based on real-world dictionaries~\cite{MikolovLS13, joulin-etal-2018-loss}. 
For the latter method, we build dictionaries based on translations from wiktionary.\footnote{\url{https://github.com/open-dsl-dict/wiktionary-dict}}
For both methods, we reduce the vocabularies to the most frequent 5k words per language and treat English as the pivot language. 

\subsubsection{Adversarial Alignment}
\label{sec:adversarialalignment}
We propose to use gradient reversal training to align embeddings from different (sub)spaces in an unsupervised way. Note that neither dictionaries nor other language resources are needed for this approach, making it applicable to zero- or low-resource scenarios.
In particular, we extend the extraction model $C$ with a discriminator $D$.
Both model parts are trained alternately in a multi-task fashion. 
The \emph{feature extractor} $F$ is shared among them and consists of the embedding layer $E$, followed by a non-linear mapping:
$F(x) = \tanh(W^\top E(x))$
with $x$ being the current word, $W \in \mathbb{R}^{S \times S}$ and $S$ being the embedding dimensionality.

The \emph{discriminator} $D$ is a multinomial non-linear classifier consisting of one hidden layer with ReLU activation \cite{hahnloser-etal-2000-digital}:\\
$D(x) = \mbox{softmax}(T^\top \mbox{ReLU}(V^\top F(x)))$
with $V \in \mathbb{R}^{S \times H}$, $T \in \mathbb{R}^{H \times O}$, $H$ being a hyperparameter and $O$ the number of different languages.

In total, we distinguish three sets of parameters: 
$\theta_C$: the parameters of the downstream classification model (i.e., the temporal tagger),
$\theta_D$: the parameters of the discriminator,
and $\theta_F$: the parameters of the feature extractor.
The loss functions of the temporal tagger $L_C$ and of the discriminator $L_D$ are cross-entropy loss functions.
While $\theta_C$ and $\theta_D$ are updated using standard gradient descent, gradient reversal training updates $\theta_F$ as follows:
\begin{align}
\theta_F &= \theta_F - \eta (\frac{\partial L_C}{\partial \theta_F} - \lambda \frac{\partial L_D}{\partial \theta_F})
\label{eq:featureupdate}
\end{align}
with $\eta$ being the learning rate and $\lambda$ a hyperparameter to control the discriminator influence.
Thus, $\theta_F$ is updated in the opposite direction of the gradients from the discriminator loss, making the discriminator an adversary.
With this, the discriminator is optimized for predicting the correct origin language of a given sentence, but at the same time the feature extractor gets updated with gradient reversal, such that the language detection becomes harder and the discriminator cannot easily distinguish the word representations from different languages.

\begin{table}
    \footnotesize
    \centering
    \begin{tabular}{l|ccc} 
        \toprule
        Dataset & Train & Dev & Test \\
        \midrule
        English (EN)    & 3,461/\emph{1,456} & 420/\emph{164} & 354/\emph{202} \\
        Spanish (ES)    & 1,705/\emph{972} & 189/\emph{122} & 332/\emph{199} \\
        Portuguese (PT) & 3,501/\emph{948} & 389/\emph{100} & 481/\emph{172} \\
        \midrule
        French (FR)     & - & - & 708/\emph{424} \\
        German (DE)     & - & - & 2,666/\emph{500} \\
        Catalan (CA)    & - & - & 1,944/\emph{1389} \\
        Basque  (EU)    & - & - & 163/\emph{123} \\
        \bottomrule
    \end{tabular}
    \caption{Number of sentences / \emph{temporal expressions} per corpus. The lower part is only used for evaluation.}
    \label{tab:datasets}
\end{table}

\newcommand{\ssig}{$^\dagger$\xspace}

\begin{table*}
    \footnotesize
    \centering
    \begin{tabular}{lr|c|cccc|cc}
        \toprule        
        & & & \multicolumn{4}{c|}{\fastText} & \multicolumn{2}{c}{BERT} \\
        \multirow{2}{*}{Task} & \multirow{2}{*}{Metric} 
        & \begin{tabular}[c]{@{}c@{}}HeidelTime\end{tabular}
        & \begin{tabular}[c]{@{}c@{}}unaligned\end{tabular} 
        & \begin{tabular}[c]{@{}c@{}}aligned \\ w/o Dict.\end{tabular} 
        & \begin{tabular}[c]{@{}c@{}}aligned \\ w/ Dict\end{tabular} 
        & \begin{tabular}[c]{@{}c@{}}aligned \\ w/ AT\end{tabular} 
        & \begin{tabular}[c]{@{}c@{}}unaligned\end{tabular} 
        & \begin{tabular}[c]{@{}c@{}}aligned \\ w/ AT\end{tabular} \\
        
        \midrule
        \multirow{3}{*}{\begin{tabular}[c]{@{}l@{}}EN\end{tabular}} 
        & strict  & \textbf{81.78} & 68.36 & 69.10 & 70.80 & 75.63 \ssig & 73.09 & 74.80 \ssig \\
        & relaxed & \textbf{90.71} & 79.14 & 79.03 & 81.21 & 82.03 \ssig & 84.34 & 86.61 \ssig\\
        & type    & \textbf{83.27} & 72.13 & 72.18 & 73.32 & 72.85 \ssig & 75.50 & 79.53 \ssig \\ 
        \midrule
        \multirow{3}{*}{\begin{tabular}[c]{@{}l@{}}ES\end{tabular}} 
        & strict  & \textbf{85.87} & 75.67 & 76.53 & 77.44 & 79.64 \ssig & 79.11 & 79.55 \\ 
        & relaxed & \textbf{90.13} & 82.43 & 82.45 & 82.47 & 84.46 \ssig & 84.12 & 85.71 \\ 
        & type    & \textbf{87.47} & 78.07 & 78.46 & 78.24 & 80.88 \ssig & 80.22 & 80.11 \\ 
        \midrule
        \multirow{3}{*}{\begin{tabular}[c]{@{}l@{}}PT\end{tabular}} 
        & strict  & 71.59 & 70.36 & 70.20 & 70.48 & 72.41 & 74.52 & \textbf{75.47} \\  
        & relaxed & 81.18 & 76.77 & 75.86 & 76.29 & 78.15 & 80.75 & \textbf{81.51} \\ 
        & type    & \textbf{76.75} & 72.29 & 71.50 & 72.26 & 73.84 & 75.47 & 76.23 \\ 
        \bottomrule
    \end{tabular}
    \caption{Results for multilingual models trained on English, Spanish and Portuguese data jointly. 
    {\ssig}highlights aligned models with statistical significant differences to the unaligned model (paired permutation test, p=0.05). }
    \label{tab:results_seen}
\end{table*}

\section{Experiments and Results}
\subsection{Evaluation Metrics and Datasets}
For evaluation, we use the TempEval3 evaluation script and report strict and relaxed extraction $F_1$ score for complete and partial overlap to gold standard annotations, respectively. We also report the type $F_1$ score for the classification into the four temporal types: Date, Time, Duration, and Set.

Our multilingual models are trained using
the Portuguese TimeBank~\cite{costa-branco-2012-timebankpt} and  TempEval3~\cite{uzzaman-etal-2013-semeval} for Spanish and English (TimeBank subset).
To demonstrate that our model is able to generalize to unseen languages, we perform tests using the French~\cite{bittar-etal-2011-french}, Catalan~\cite{sauri-catalan} and Basque~\cite{altuna2016adapting} TimeBanks and the Zeit subset of the German KRAUTS corpus \cite{strotgen-etal-2018-krauts}. 
Corpus statistics are shown in Table~\ref{tab:datasets}.

\subsection{Hyperparameters and Model Training}
We use the AdamW optimizer~\cite{loshchilov2018decoupled} with a learning rate of \num{1e-5} for the BiLSTM-CRF model part and \num{1e-6} for BERT.
The model is trained for a maximum of 50 epochs using early stopping on the development set. The BiLSTM has a hidden size of 128 units per direction. The labels are encoded in the IOB2 format. For regularization, we apply dropout with a rate of 10\% after the input embeddings.  
The discriminator for adversarial training has a hidden size $H$ of 100 units and is trained after every $10^{th}$ batch of the sequence tagger with $\lambda$ set to 0.001.

\subsection{Results}
The results for the multilingual experiments are shown in Table~\ref{tab:results_seen}.
We trained three models with different random seeds and report the performance of the model with median performance on the combined development set of all languages. 
Current state of the art for English~\cite{lee-etal-2014-context} achieves 83.1/91.4/85.4 for strict/relaxed/type $F_1$. 
However, this is a monolingual model that can only be applied to English.

The effects of aligning \fastText embeddings are clearly visible in Table~\ref{tab:results_seen}. The supervised alignment using a dictionary is always superior compared to the unsupervised alignment without a dictionary or the unaligned embeddings. Our proposed adversarial alignment (w/ AT) leads to the best results across languages. The performance of BERT is close to the best \fastText model.\footnote{Additional experiments with the multilingual XLM model~\cite{lample2019cross} trained on 100 languages led to similar results as the multilingual BERT model.}
Aligning BERT with adversarial training also increases performance. The improvements are smaller compared to \fastText but still statistically significant for English. 

\begin{table}
    \footnotesize
    \centering
	\setlength\tabcolsep{4.5pt}
    \begin{tabular}{lr|c|cc}
        \toprule
        & & & \multicolumn{2}{c}{BERT} \\
        \multirow{2}{*}{Task} & \multirow{2}{*}{Metric} 
        & \begin{tabular}[c]{@{}c@{}}HeidelTime\\-Auto  \end{tabular}
        & \begin{tabular}[c]{@{}c@{}}unalign. \end{tabular} 
        & \begin{tabular}[c]{@{}c@{}}aligned \\ w/ AT\end{tabular} \\
        \midrule
        \multirow{3}{*}{\begin{tabular}[c]{@{}l@{}}FR\end{tabular}} 
        & strict  & 52.35 & 60.12 & \textbf{62.58} \\ 
        & relaxed & 72.02 & 74.23 & \textbf{75.46} \\ 
        & type    & \textbf{68.70} & 61.96 & 62.07 \\ 
        \midrule
        \multirow{3}{*}{\begin{tabular}[c]{@{}l@{}}DE\end{tabular}} 
        & strict  & 38.87 & 63.34 & \textbf{66.53} \\ 
        & relaxed & 52.11 & 76.51 & \textbf{77.82} \\
        & type    & 50.15 & 66.95 & \textbf{69.04} \\ 
        \midrule
        \multirow{3}{*}{\begin{tabular}[c]{@{}l@{}}CA\end{tabular}} 
        & strict  & 28.11 & 63.24 & \textbf{64.21} \\
        & relaxed & 62.81 & 74.95 & \textbf{77.00} \\
        & type    & 60.84 & 65.66 & \textbf{67.85}\\
        \midrule
        \multirow{3}{*}{\begin{tabular}[c]{@{}l@{}}EU\end{tabular}} 
        & strict  & 22.54 & 43.96 & \textbf{47.87} \\
        & relaxed & 26.76 & 61.54 & \textbf{63.83} \\
        & type    & 23.94 & 57.14 & \textbf{58.51} \\
        \bottomrule
    \end{tabular}
    \caption{Results for the unsupervised cross-lingual setting. We compare to HeidelTime with automatically generated resources, which resembles a similar setting.}
    \label{tab:results_unseen}
	\setlength\tabcolsep{6pt}
\end{table}

Table~\ref{tab:results_unseen} provides transfer results of the models with BERT embeddings to languages without labeled training data.\footnote{The results of the \fastText models were considerably lower for cross-lingual transfer. }
In particular, the model using the Wikipedia data for training the discriminator is effective in generalizing to languages without training resources for temporal expression extraction, as these languages are also aligned during model training. 
It outperforms the state-of-the-art HeidelTime models by a large margin.
The impressive performance of the multilingual BERT in the cross-lingual setting can be explained by the fact that the model has seen many sentences in our target languages
during the pre-training phase, which can now be effectively leveraged in this new setting.

\subsection{Analysis}
\begin{figure}
    \centering
    \begin{subfigure}[t]{0.23\textwidth}
        \centering
        \includegraphics[width=1.0\textwidth]{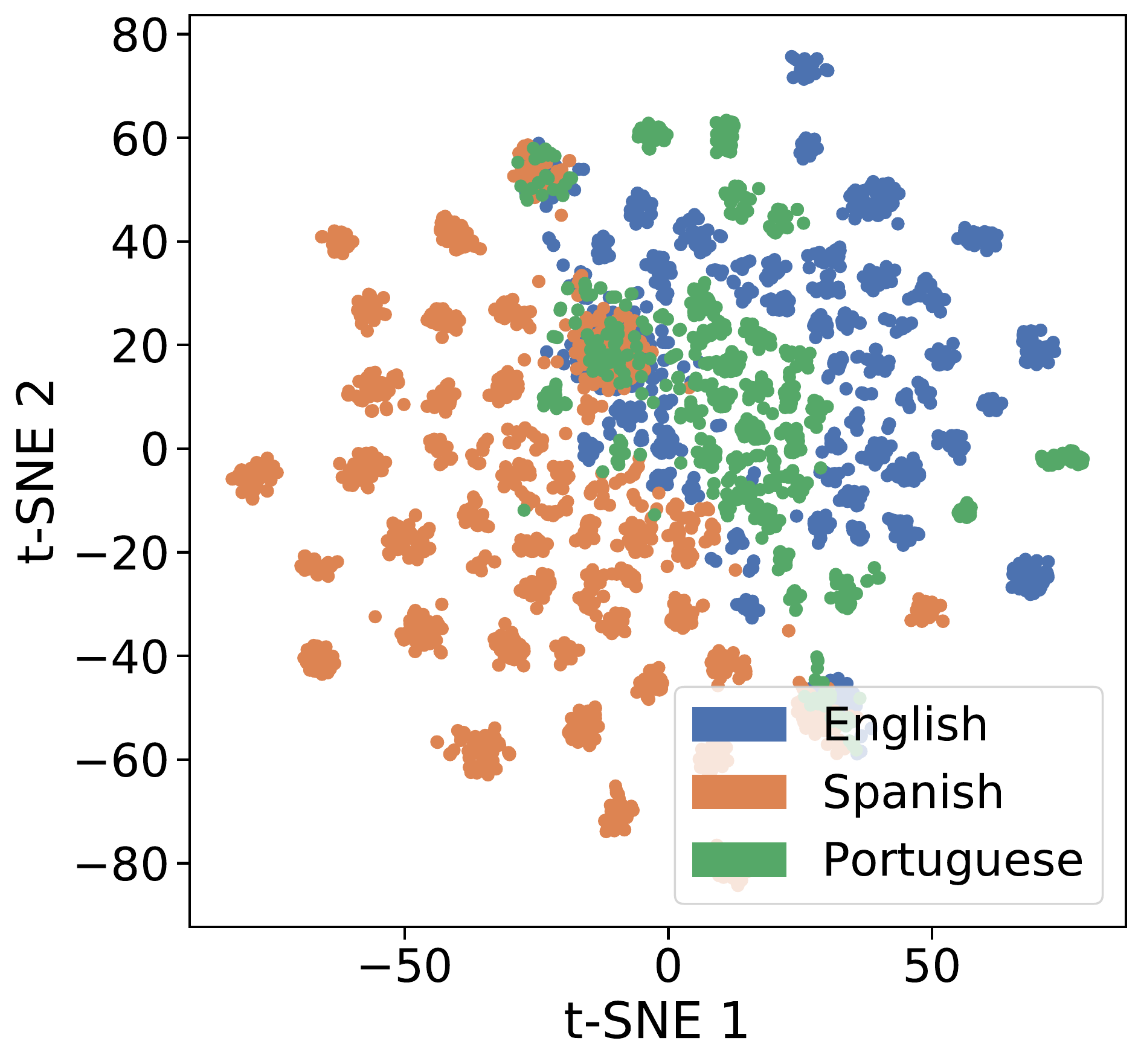}
        \caption{without training.}
        \label{fig:space_start}
    \end{subfigure}
    ~
    \begin{subfigure}[t]{0.23\textwidth}
        \centering
        \includegraphics[width=1.0\textwidth]{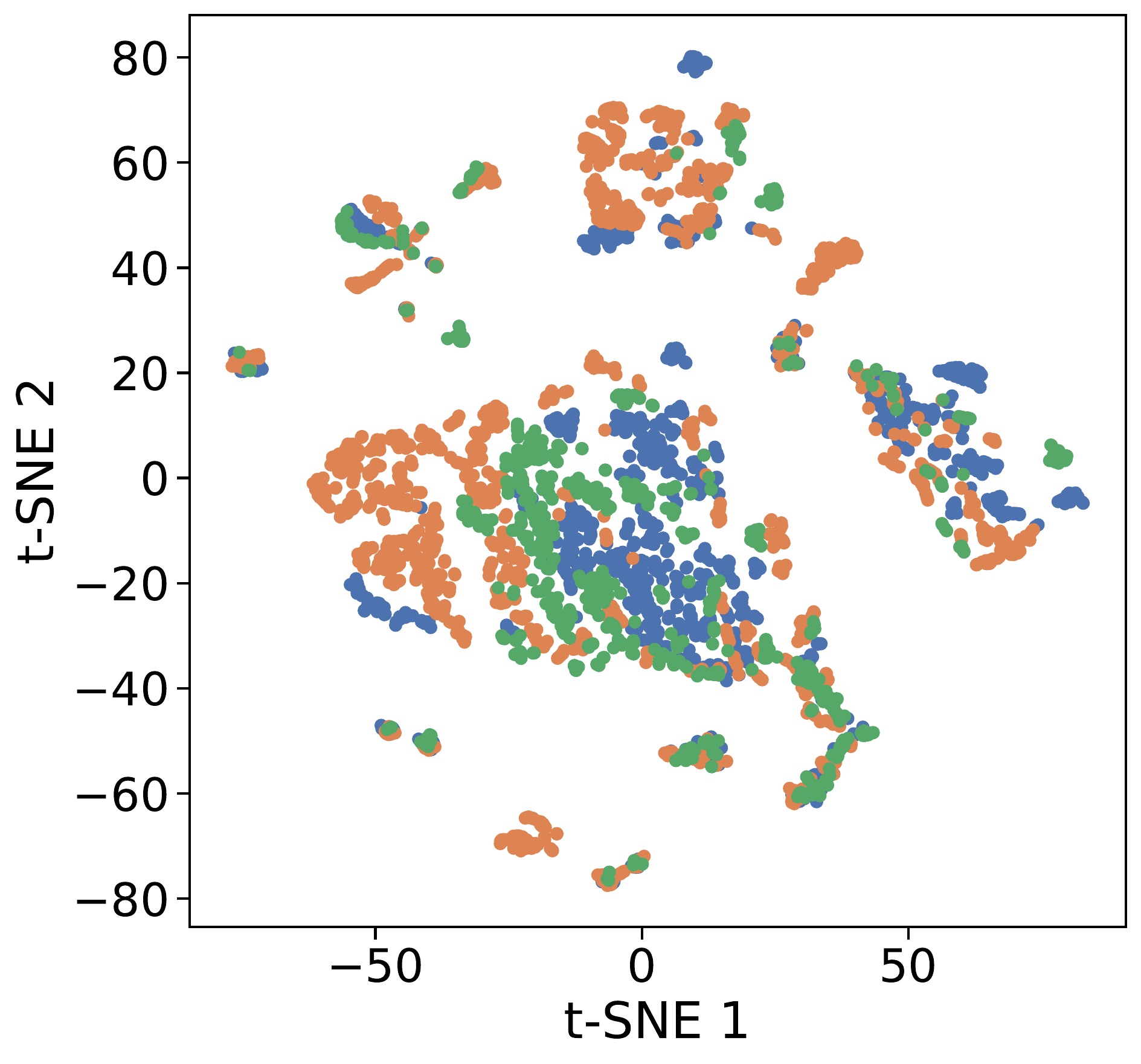}
        \caption{after joint temporal extraction and adversarial training.}
        \label{fig:space_adv}
    \end{subfigure}
    \caption{t-SNE plots of the last BERT layer without any training (left)
    and after training (right). }
    \label{fig:embedding_space}
\end{figure}

The embedding spaces of BERT before and after aligning are shown in Figure~\ref{fig:embedding_space}. The left sub-figure presents the original BERT embeddings without any fine-tuning. In this visualization, clear clusters for each language exist. 
After fine-tuning on multilingual temporal expression extraction 
and adversarial alignment (right sub-figure) the clusters for each language mostly disappear. 

\section{Conclusion}
In this paper, we investigated how a multilingual neural model with \fastText or BERT embeddings can be used to extract temporal expressions from text. 
We investigated adversarial training for creating
multilingual embedding spaces.
The model can effectively be transferred to unseen languages in a cross-lingual setting and outperforms a state-of-the-art model by a large margin.

\section*{Acknowledgments}
We would like to thank the members of the BCAI NLP\&KRR research group and the anonymous reviewers for their helpful comments.

\bibliographystyle{acl_natbib}
\bibliography{references}

\end{document}